\documentclass{article}

\usepackage[final]{neurips_2020}

\usepackage[utf8]{inputenc} 
\usepackage[T1]{fontenc}    
\usepackage[colorlinks,citecolor=black,linkcolor=black]{hyperref}
\usepackage{url}            
\usepackage{booktabs}       
\usepackage{amsfonts}       
\usepackage{nicefrac}       
\usepackage{microtype}      
\usepackage{multirow}
\usepackage{amsmath}
\usepackage{graphicx}
\usepackage[caption=false]{subfig}
\usepackage{color}
\usepackage{wrapfig}
\usepackage[linesnumbered,ruled,vlined]{algorithm2e}
\usepackage{algpseudocode}
\graphicspath{{./graph/},{./}}
\DeclareGraphicsExtensions{.pdf,.jpg,.png}

\ifdefined \GramaCheck
\newcommand{\CheckRmv}[1]{}
\newcommand{\figref}[1]{Figure}
\newcommand{\secref}[1]{Section}
\newcommand{\tabref}[1]{Table}
\newcommand{\algorref}[1]{Algorithm}
\renewcommand{\eqref}[1]{Equation}
\else
\newcommand{\CheckRmv}[1]{#1}
\newcommand{\figref}[1]{Fig.~\ref{#1}}
\newcommand{\secref}[1]{Sec.~\ref{#1}}
\newcommand{\tabref}[1]{Tab.~\ref{#1}}
\newcommand{\algorref}[1]{Algorithm.~\ref{#1}}
\renewcommand{\eqref}[1]{Eq.~\ref{#1}}
\fi

\title{Generalized Zero-Shot Learning \\ via VAE-Conditioned Generative Flow}

\author{%
	Yu-Chao Gu\textsuperscript{\rm 1}, Le Zhang\textsuperscript{\rm 2}, Yun Liu\textsuperscript{\rm 1}, 
		Shao-Ping Lu\textsuperscript{\rm 1}, Ming-Ming Cheng\textsuperscript{\rm 1}\\
	\textsuperscript{\rm 1}TKLNDST, CS, Nankai University, China\\
	\textsuperscript{\rm 2}Agency for Science Technology and Research (A*STAR), Singapore
}

\begin{document}
	
	\maketitle
	
	\begin{abstract}
		Generalized zero-shot learning (GZSL) aims to recognize both seen 
		and unseen classes by transferring knowledge from semantic descriptions to visual representations. 
		Recent generative methods formulate GZSL as a missing data problem, which mainly adopts GANs or VAEs to generate visual features for unseen classes.
		However, GANs often suffer from instability, and VAEs can only optimize 
		the lower bound on the log-likelihood of observed data.
		To overcome the above limitations, we resort to generative flows, a family of generative models with the advantage of accurate likelihood estimation. More specifically, we propose a conditional version of generative flows for GZSL, \textit{i.e.}, VAE-Conditioned Generative Flow (VAE-cFlow). 
		By using VAE, the semantic descriptions are firstly encoded into tractable latent distributions, conditioned on that the generative flow optimizes the exact log-likelihood of the observed visual features.
		We ensure the conditional latent distribution to be both semantic meaningful and inter-class discriminative by i) adopting the VAE reconstruction objective, ii) releasing the zero-mean constraint in VAE posterior regularization, and iii) adding a classification regularization on the latent variables.
		Our method achieves state-of-the-art GZSL results on five well-known benchmark datasets, especially for the significant improvement in the large-scale setting. 
		Code is released at \url{https://github.com/guyuchao/VAE-cFlow-ZSL}.
	\end{abstract}

	\section{Introduction}
	Balanced datasets are essential for building an accurate vision 
	recognition system in the deep learning era. 
	However, the number of object classes in real-world could be  unbounded, of which only a small subset could be collected and annotated, leaving most object classes untouched. 
	Zero-shot learning (ZSL) is proposed to tackle this 
	challenging problem. 
	It aims to recognize unseen classes that are not included in the 
	training phase. 
	A more realistic setting is generalized zero-shot learning (GZSL), 
	in which both the seen and unseen classes can be involved 
	during testing.
	
	ZSL mainly works by utilizing the semantic descriptions, such as 
	the semantic attributes and word embeddings. Since the semantic space contains both seen and unseen classes, one straightforward way is to learn a visual-semantic mapping \cite{norouzi2013zero,lampert2013attribute} and perform classification tasks in the semantic space.
	The visual-semantic mapping is designed to model linear compatibility
	\cite{akata2015label} and non-linear compatibility \cite{xian2016latent} 
	between visual features and semantic descriptions. Because such mapping functions are trained on seen classes, the mapping of unseen features will be biased to seen classes. Hence most of ZSL methods fail for the more realistic GZSL.
	
	Apart from the above solutions, some efforts on generative methods convert GZSL into the traditional classification problem by generating sufficient visual representations for unseen classes. 
	The performance of GZSL depends on how well the generative methods 
	model the conditional probability distribution of visual features \cite{xian2018feature}. 
	Notable works within this family are based on the Generative 
	Adversarial Networks (GANs)
	\cite{bucher2017generating,xian2018feature,felix2018multi,zhu2018generative} 
	and Variational Auto-Encoders (VAEs) \cite{yu2019zero,schonfeld2019generalized,mishra2018generative,wang2018zero}. 
	The GAN-based methods adopt a conditional generator to transform random noise and conditional information (\textit{i.e.}, semantic descriptions) into visual features, where a discriminator 
	is used to indicate the optimizing direction of the generator. 
	Such adversarial optimization has the risk of modal collapse
	\cite{dumoulin2016adversarially,metz2016unrolled,che2016mode} 
	if the power of the generator and discriminator is unbalanced. 
	The VAE-based methods, which optimize the evidence lower bound on the log-likelihood of observed data, are more stable to train \cite{larsen2015autoencoding}. However, restricted by their objective designs, the generation ability of VAE-based methods is inferior to that 
	of GAN-based methods \cite{huang2018introvae,zhao2019infovae}.
	
	To overcome the above limitations, we resort to the \textit{generative flows} \cite{dinh2014nice,kobyzev2019normalizing}, a family of generative models for image generation and density estimation. The generative flows consist of a sequence of invertible transformations that transform a sample from the simple distribution with tractable density (\textit{e.g.}, multivariate Gaussian) into a complex distribution.
	Based on \textit{the change of variable theory}, the log-likelihood of the sample under complex distribution is tractable by transforming it back to the simple distribution and computing the sum of i) the log-likelihood of the reversely-transformed sample under the simple distribution and ii) the change in the volume induced by the transformation sequence. Different from VAE-based methods which can only optimize the lower bound on the log-likelihood of observed visual features, the generative flow directly optimizes the accurate log-likelihood objective. However, due to its difficulties in the conditional generation, it is not straightforward to apply the generative flow into GZSL.
	
	To address this problem, we propose a conditional version of the generative flow for GZSL, namely, VAE-Conditioned Generative Flow (VAE-cFlow). VAE-cFlow encodes the conditional information, \textit{i.e.}, semantic descriptions, into a tractable latent distribution which is both semantic meaningful and inter-class discriminative. This is achieved by using a combination of a reconstruction loss and a posterior regularization. The former guarantees the encoding can be decoded back to the original input, which essentially forces the conditional information to be semantic meaningful. The latter encourages the inter-class discrimination of the latent distribution by replacing the zero-mean constraint of the VAE posterior regularization with classification regularization on the latent variable.  To the best of our knowledge, this is the first work to generalize 
	the generative flow to the conditional feature generation for GZSL. 
	Our contributions are summarized as follows:
	i) We propose a conditional probabilistic generative model VAE-cFlow for GZSL, with the advantage of constructing semantic meaningful conditional latent distribution and accurate likelihood optimization.
	ii) We conduct extensive experiments on AwA1 \cite{lampert2013attribute}, AwA2 \cite{xian2017zero}, CUB \cite{wah2011caltech}, SUN \cite{patterson2012sun} and ImageNet21K \cite{deng2009imagenet} datasets and achieve state-of-the-art results. 
	More specifically, we achieve $3.9\%$ and $3.1\%$ improvement on the M500 and M1K splits of ImageNet21K, respectively.
	iii) We demonstrate that the proposed method can generate features with proper intra-class variation, which is essential to improve GZSL performance.

	\section{Related Work}
	\subsection{Zero-shot Learning}
	Zero-shot learning (ZSL) aims to tackle the extremely class-imbalanced 
	problem, in which no visual examples for unseen classes are available 
	for training. 
	A more challenging yet realistic setting is called generalized zero-shot 
	learning (GZSL) where both the seen and unseen classes could be 
	involved in the testing phase. 
	Since there are no training examples for unseen class, semantic 
	descriptions, such as semantic attributes and word embeddings, 
	are usually required to achieve ZSL/GZSL.
	Prior works are mainly based on learning a mapping function 
	between the visual representations and the semantic descriptions. 
	We refer readers to \cite{xian2017zero} for a comprehensive survey.
	
	Here, we focus on recent generative GZSL methods. \cite{bucher2017generating,xian2018feature,zhu2018generative,arjovsky2017wasserstein} adopt conditional GAN to generate unseen visual representations. Among them, GAZSL \cite{zhu2018generative} and FGZSL \cite{xian2018feature} add a classification regularization on synthetic features to ensure their inter-class discrimination. MCGZSL \cite{arjovsky2017wasserstein} proposes a multi-modal cycle-consistent regularization to ensure synthetic visual features to be semantic meaningful. Another family of generative methods \cite{mishra2018generative,wang2018zero,yu2019zero,schonfeld2019generalized} for GZSL is built upon VAE \cite{kingma2013auto}. CVAE-ZSL \cite{mishra2018generative} and VZSL \cite{wang2018zero} apply conditional VAE to synthesize unseen features. SGAL \cite{yu2019zero} adopts the expectation-maximization 
	(EM) algorithm, iteratively learning to generate and classify based on multi-modal VAE.
	CADA-VAE \cite{schonfeld2019generalized} aligns visual and semantic distributions to construct latent features that contain the essential multi-modal information.
	
	However, GAN-based methods usually suffer from training instability
	\cite{dumoulin2016adversarially,metz2016unrolled,che2016mode}, 
	while the generation ability of VAE-based methods is inferior to 
	the GAN-based methods \cite{huang2018introvae,zhao2019infovae}. 
	Recently, f-VAEGAN-D2 \cite{xian2019f} combines the strength of GANs and VAEs. ABP-ZSL \cite{zhu2019learning} stabilizes the training procedure by ablating the discriminator of GAN and optimizes the generator by EM and alternating backpropagation algorithm. Different from previous methods, we construct VAE-cFlow based on the generative flow, which can stably optimize the accurate log-likelihood objective. 
	
	\subsection{Generative Flow}
	Generative flows \cite{kobyzev2019normalizing} are a family of generative methods with tractable distribution, which can optimize the exact log-likelihood of observed data. Through a sequence of invertible functions transforming a sample from a simple distribution to a complex distribution, the density of the sample under the complex distribution can be exactly computed based on \textit{the change of variable theory}. 
	Previous generative flows \cite{dinh2014nice,dinh2016density,kingma2018glow}
	have limited ability to model the conditional distribution. Recently,
	Liu \textit{et al.} \cite{liu2019conditional} proposed an adversarial strategy to encode conditional information into the latent space.
	Sun \textit{et al.} \cite{sun2019dual} leveraged a pair of flows 
	for conditional generation on each prior distribution. However, the conditional latent distribution for GZSL is required to be semantic meaningful and inter-class discriminative, which is not guaranteed in previous works. In this paper, we demonstrate such a conditional latent space can be constructed by modifying the VAE objective. 
	
	\CheckRmv{
		\begin{figure*}[t]
			\centering
			\includegraphics[width=.98\linewidth]{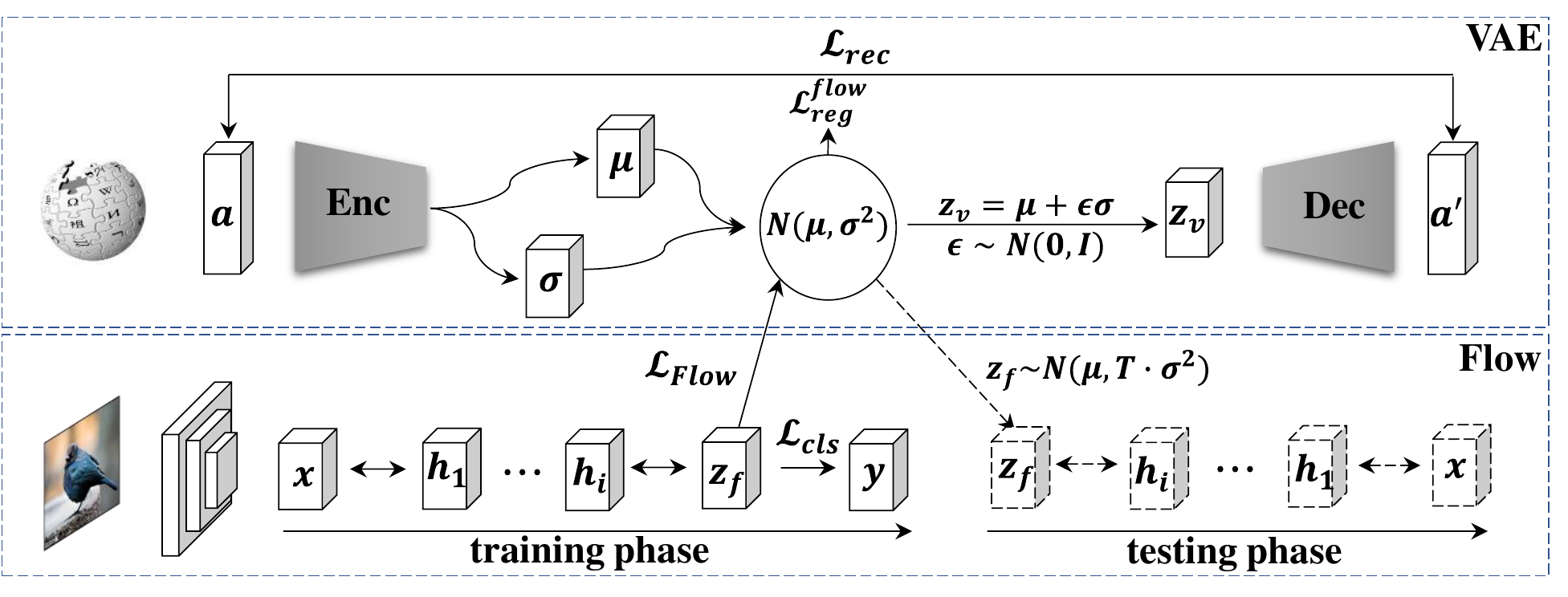}
			\vspace{-.1in}
			\caption{Illustration of the proposed VAE-cFlow. 
				The semantic description is encoded into a latent 
				Gaussian distribution, conditioned on that the generative flow optimizes the log-likelihood objective, \textit{i.e.},
				$\mathcal{L}_{flow}$. 
				The trained generative flow is inverted to generate visual features in the testing phase.}
			\label{fig:network}
			\vspace{-.1in}
	\end{figure*}}

	\section{Method}
	In this section, we first introduce the notations for zero-shot 
	learning in \secref{sec:notation}. 
	In \secref{sec:flow}, we describe how to adapt the generative flow 
	to conditional feature generation 
	by introducing the semantic-specific latent distribution. 
	In \secref{sec:vae}, we show the semantic-specific latent distribution can be obtained 
	by modifying the posterior regularization in VAE objective and adding the classification
	regularization. 
	In \secref{sec:obejctive}, we describe the overall objective of 
	VAE-cFlow and the training and testing procedures.

	\subsection{Preliminary} \label{sec:notation}
	In zero-shot learning, we have access to a training dataset
	$\mathcal{D}_{s}=\begin{Bmatrix}(x_{i},a_{i},y_{i})\end{Bmatrix}_{i=1}^{n}$, 
	where $x_{i}\in \mathcal{X}_{s}$ and $a_{i}\in \mathcal{A}_{s}$ 
	stand for visual features and semantic descriptions (\textit{e.g.} 
	the word embeddings or semantic attributes) corresponding to the seen class 
	$y_{i}\in \mathcal{Y}_{s}$, respectively. 
	In the same way, we denote $\mathcal{Y}_{u}$, $\mathcal{X}_{u}$, and $\mathcal{A}_{u}$ as unseen classes and their corresponding features 
	and semantic descriptions, respectively. 
	Given the training set
	$\begin{Bmatrix}\mathcal{X}_{s},\mathcal{A}_{s},\mathcal{Y}_{s}\end{Bmatrix}$
	and $\begin{Bmatrix}\mathcal{A}_{u},\mathcal{Y}_{u}\end{Bmatrix}$, 
	the goal of zero-shot learning is to recognize unseen features
	$\mathcal{X}_{u}$ in the testing phase.
	For generalized zero-shot learning, the search space contains both 
	the seen classes $\mathcal{Y}_{s}$ and unseen classes $\mathcal{Y}_{u}$.

	\subsection{Generative Flow Feature Generation} \label{sec:flow}
	Given the training dataset
	$\mathcal{D}_{s}=\begin{Bmatrix}(x_{i},a_{i},y_{i})\end{Bmatrix}_{i=1}^{n}$ 
	observed from the semantic-conditional  visual distribution 
	$x_{i}|a_{i}\sim p^{*}(x_{i}|a_{i})$, 
	we aim to learn a model $p_{\theta}$ parameterized by $\theta$ to maximize 
	the likelihood of the observed visual features, which is equivalent to minimizing 
	the following negative log-likelihood objective:
	\begin{equation}
	\label{eq:likelihood}
	\mathcal{L}_{flow}= \sum_{i=1}^{n} -\log p_{\theta}(x_{i}|a_{i}).
	\end{equation}
	We denote an invertible function as $z_f=f_{\theta}(x)$ that 
	transforms visual features into latent variables. 
	The latent variable $z_f$ has a tractable probability density function $p_{\theta}(z_f|a)$ conditioned 
	on the semantic information $a$, \textit{i.e.}, the multivariate Gaussian
	distribution with diagonal covariance matrix
	$p_{\theta}(z_f|a)=\mathcal{N}(z_f;\mu(a),\sigma^{2}(a)\mathbf{I})$.
	Based on \textit{the change of variable theory},
	the conditional probability density of visual feature $x$ can be obtained by calculating
	\begin{equation}
	\label{eq:changevarible}
	\log p_{\theta}(x|a)=\log p_{\theta}(z_f|a) + \log \left(\left| det(\frac{dz_f}{dx}) \right| \right),
	\end{equation}
	where the first term is the conditional probability density of the latent variable $z_f$. 
	The second term is the logarithm of the absolute value of the Jacobian of $f_\theta$, measuring the change in the volume induced by the invertible transformation. 
	While we could have many choices for the transformation functions, 
	in practice we mainly consider those functions that are easy to invert and
	compute the determinant of their Jacobian. 
	Especially, if the Jacobian of the transformations is triangular, the determinant of those transformations has a simple form, \textit{i.e.}, the product of the diagonals of the Jacobian matrix. 
	We continue by introducing two basic components that satisfy this property, 
	\textit{i.e.}, the additive coupling layer \cite{dinh2014nice} and actnorm \cite{kingma2018glow}.

	\paragraph{Additive coupling layer} 
	Considering a partition of the input $x$ along channels into two 
	parts, \textit{i.e.}, $x_{a}$ and $x_{b}$, the additive coupling 
	layer (left) and its inversion form (right) can be formulated as
	
	\begin{equation}
	\begin{gathered}
	\left\{
	\begin{array}{l}
	y_{a}=x_{a}+t(x_{b}), \\  y_{b}=x_{b},
	\end{array}
	\right.
	\end{gathered}  \qquad
	\begin{gathered}
	\left\{
	\begin{array}{l}
	x_{a}=y_{a}-t(y_{b}), \\  x_{b}=y_{b}.
	\end{array}
	\right.
	\end{gathered}    
	\end{equation}
	
	The additive coupling layer has the unit determinant of Jacobin and 
	a simple inverse form.
	The inversion of an additive coupling layer do not involve the inversion of function $t$, and $t$ can thus be arbitrarily complex. In practice, it is usually modeled by neural networks. 
	
	\paragraph{Actnorm}
	Actnorm is proposed to replace Batchnorm \cite{ioffe2015batch} 
	in flow-based models. 
	Actnorm performs a learnable affine transformation using the scale 
	and bias parameters per channel. 
	The actnorm (left) and its inversion form (right) can be formulated as
	\begin{equation}
	y=s\odot x+b, \qquad
	x=(y-b)/s.
	\end{equation}
	The determinant of its Jacobian is the product of the scales $s$, \textit{i.e.},
	$hw{\sum(|s|)}$, where $h$ and $w$ mean the spatial dimensions of $x$.
	
	Once trained, we can invert $f_{\theta}$ to obtain the generator 
	function $g_{\theta}(z_f)=f^{-1}_{\theta}(x)$.  
	The generation process can be defined as
	\begin{equation}
	\label{eq:gene}
	z_f\sim p_{\theta}(z_f|a), \qquad x=g_{\theta}(z_f),
	\end{equation}
	where $p_{\theta}(z_f|a)$ is the conditional probability density function that can be obtained by VAE.

	\subsection{Conditional Distribution Encoded via VAE} \label{sec:vae}
	\eqref{eq:changevarible} shows that optimizing the log-likelihood objective of generative flow is based on the conditional distribution of the latent variable $p_{\theta}(z_f|a)=\mathcal{N}(z_f;\mu(a),\sigma^{2}(a)\mathbf{I})$.
	The parameters of latent Gaussian can be encoded by VAE.
	Formally, by introducing an encoder $q_{\varphi}(z_v|a)$ and a decoder 
	$p_{\omega}(a|z_v)$, the optimization objective of VAE can 
	be defined as
	\begin{equation}
	\label{eq:VAE}
	\mathcal{L}_{\textit{\text{VAE}}}=-\mathbb{E}_{q_{\varphi}(z_v|a)}[\log p_{\omega}(a|z_v)]+KL(q_{\varphi}(z_v|a)||(p(z_v))=\mathcal{L}_{rec}+\mathcal{L}_{reg}^{origin}.
	\end{equation}
	The first term is the reconstruction error between the input semantic embedding and the embedding decoded back from latent $z_v$, which is also called variational lower bound. The second term is a posterior regularization, encouraging the latent distribution $q_{\varphi}(z_v|a)$ to be close to the prior distribution $p(z_v)=\mathcal{N}(z_v;0,\mathbf{I})$. The conditional latent distribution for generative flow is required to be both semantic meaningful and inter-class discriminative. The reconstruction error ensures VAE to encode semantic meaningful information that can be decoded back to the input. The original posterior regularization \cite{kingma2013auto} computes the Kullback-Leibler divergence between the posterior and prior distributions, which can be calculated without estimation as
	\begin{equation}
	\mathcal{L}_{reg}^{origin}=\frac{1}{2}\sum_{i=1}^{d}\mu_{i}^{2}+\frac{1}{2}\sum_{i=1}^{d}(\sigma_{i}^{2}-\log\sigma_{i}^{2}-1)=\mathcal{L}_{\mu}+\mathcal{L}_{\sigma},
	\end{equation}
	where $\mu_{i}$ and $\sigma_{i}^{2}$ are obtained from the encoder 
	of VAE, and $d$ is the dimension of the latent Gaussian distribution. The original posterior regularization is composed of the zero-mean constraint ($\mathcal{L}_{\mu}$) and the unit variance constraint ($\mathcal{L}_{\sigma}$). The zero-mean constraint makes the latent distribution less inter-class discriminative. Hence we release the zero-mean constraint to implicitly encourage the latent distribution from different classes to have different means, which can be formulated as
	\begin{equation}
	\mathcal{L}_{reg}^{flow}=\mathcal{L}_{\sigma}=\sum_{i=1}^{d}(\sigma_{i}^{2}-\log\sigma_{i}^{2}-1).
	\end{equation}
	
	We add a classification regularization for the latent variable $z_v$, 
	further encouraging the separation of the latent distribution from different classes. We apply softmax activation on $z_v$ to get the probability distribution  $p_i$, and then the classification regularization can be written as
	\begin{equation} \label{eq:loss_cls}
	\mathcal{L}_{cls}=-\sum_{j\in \mathcal{Y}_{s}}y_{i}^{(j)}\log p_{i}^{(j)},
	\end{equation}
	where $y_{i}$ is the one-hot label associated with the given semantic 
	attribute $a_{i}$. 
	
	\CheckRmv{
		\begin{algorithm}[!tb]
			\label{alg:train}
			\caption{VAE-cFlow training algorithm}
			\KwIn{The seen dataset $\mathcal{D}=\{(x_i, a_{i}, y_{i})\}_{i=1}^n$, 
				the number of training iterations $N_{iter}$}
			\KwOut{The learned parameters $\theta, \omega, \varphi$}
			Initialize $\theta, \omega, \varphi$\;
			\For{$iter = 1,\dots, N_{iter}$}
			{
				$x_{i}, a_{i}, y_{i}\leftarrow$ Randomly sample a data-point 
				from $\mathcal{D}$\;
				$\mu(a_{i}), \sigma^{2}(a_{i})\leftarrow$ $q_{\varphi}(a_{i})$, 
				encode the semantic description $a_{i}$ to Gaussian parameters\;
				$z_{v}\leftarrow$ Sample latent from $\mathcal{N}(z_v;\mu(a_{i}),\sigma^{2}(a_{i}))$ 
				using the reparameterization trick \cite{kingma2013auto}\;
				$a_{i}'\leftarrow p_{\omega}(z_{v})$, decode $z_{v}$ to the semantic description\;
				$\mathcal{L}_{rec}\leftarrow$ Measure the $l_{1}$ reconstruction error between $a_{i}$ and $a_{i}'$\;
				$\mathcal{L}_{reg}^{flow}\leftarrow$ Calculate the unit variance regularization\;
				$z_{f}\leftarrow f_{\theta}(x_{i})$, transform $x_{i}$ into a latent variable\;
				$\mathcal{L}_{flow}\leftarrow$ Calculate the log-likelihood objective of generative flow\;
				$\mathcal{L}_{cls}\leftarrow$ Calculate the Cross-Entropy loss between $z_{f}$ and $y_{i}$\;
				$g\leftarrow \nabla( \mathcal{L}_{rec}+\mathcal{L}_{reg}^{flow}+\mathcal{L}_{flow}+\mathcal{L}_{cls})$\;
				$\theta, \omega, \varphi \leftarrow$ Update parameters through gradient $g$\;
			}
			return $\theta, \omega, \varphi$\;
	\end{algorithm}}

	\subsection{Overall Objective} \label{sec:obejctive}
	The VAE-cFlow jointly optimizes the generative flow and VAE through the overall objective
	\begin{equation}
	\mathcal{L}=\mathcal{L}_{flow}+\mathcal{L}_{rec}+\mathcal{L}_{reg}^{flow}+\mathcal{L}_{cls}.
	\end{equation}
	The whole architecture of VAE-cFlow is shown in \figref{fig:network}. In training, we follow the algorithm described in \algorref{alg:train}. 
	After training, we invert the generative flow to generate visual features for unseen classes from latent, sampling from the conditional latent distribution encoded by VAE:
	\begin{equation}
	\label{eq:test}
	z\sim \mathcal{N}(z_f;\mu(a),\mathcal{T}\cdot \sigma^{2}(a)\mathbf{I}), \qquad
	x=g_{\theta}(z),
	\end{equation}
	where the sampling temperture $\mathcal{T}$ controls the intra-class variation of the latent variable and synthetic visual features. Once we obtain synthetic features for unseen classes, we train a simple linear classifier to classify both seen and unseen classes.
	
	\CheckRmv{
		\begin{table}[!tb]
			\centering
			\caption{Generalized zero-shot learning evaluation on AwA1, AwA2, CUB, and SUN datasets.}
			\label{tab:GZSLresults}
			\scriptsize
			\begin{tabular}{c|l|ccc|ccc|ccc|ccc}
				\toprule[2pt]
				& \multirow{2}*{\textbf{Methods}}& \multicolumn{3}{c|}{AwA1} & \multicolumn{3}{c|}{AwA2} &\multicolumn{3}{c|}{CUB} & \multicolumn{3}{c}{SUN} \\ 
				& & $\mathit{A}_{\mathcal{U}}$ & $\mathit{A}_{\mathcal{S}}$ & \multicolumn{1}{c|}{$\mathit{H}$} & $\mathit{A}_{\mathcal{U}}$ & $\mathit{A}_{\mathcal{S}}$ & \multicolumn{1}{c|}{$\mathit{H}$} & $\mathit{A}_{\mathcal{U}}$ & $\mathit{A}_{\mathcal{S}}$ & \multicolumn{1}{c|}{$\mathit{H}$} & $\mathit{A}_{\mathcal{U}}$ & $\mathit{A}_{\mathcal{S}}$ & $\mathit{H}$\\ \midrule[0.5pt]
				\multirow{7}{*}{\rotatebox{90}{\hspace{-1.3mm} Non-generative}} &
				IAP \cite{lampert2013attribute}  &   2.1  &  78.2  &  4.1  &  0.9   &  87.6  &  1.8  &  0.2  &  72.8  &  0.4  &  1.0  &  37.8  &  1.8  \\
				& CMT \cite{socher2013zero}   &   0.9  &  87.6  &  1.8  &  0.5   &  90.0  &  1.0  &  7.2  &  49.8  &  12.6  &  8.1  &  21.8  &  11.8  \\
				& LATEM \cite{xian2016latent} &  7.3  &  71.7  &  13.3  &  11.5   &  77.3  &  20.0  &  15.2  &  57.3  &  24.0  &  14.7  &  28.8  &  19.5  \\
				& SYNC \cite{changpinyo2016synthesized} &  8.9  &  87.3  &  16.2  &  10.0   &  90.5  &  18.0  &  11.5  &  70.9  &  19.8  &  7.9  &  43.3  &  13.4  \\
				& ALE \cite{akata2015label}   &   16.8  &  76.1  &  27.5  &  14.0   &  81.8  &  23.9  &  23.7  &  62.8  &  34.4  &  21.8  &  33.1  &  26.3  \\ 
				& ESZSL \cite{romera2015embarrassingly}  &  6.6  &  75.6   & 12.1  &  5.9   &  77.8  &  11.0   &  12.6  &  63.8   &  21.0  &  11.0  &  27.9 &  15.8  \\ 
				& DeViSE \cite{frome2013devise}   &  13.4   &  68.7   & 22.4  &  17.1  &  74.7  &  27.8  &  23.8  &  53.0  & 32.8  & 16.9 &  27.4  & 20.9  \\ \midrule[0.5pt] 
				\multirow{7}{*}{\rotatebox{90}{\hspace{-2.0mm} Generative}} &
				GAZSL \cite{zhu2018generative}   &  32.8  &  84.7  &  47.3  &  59.9  &  68.3  & 53.4  &  26.5  &  57.4  &  36.2  & 21.7  &  34.5   & 26.7 \\
				& FGZSL \cite{xian2018feature}  &  53.1  &  68.0  &  59.6   &  50.2  &  67.5   &  57.5  &  45.9   &  54.6  & 49.9      & 40.2  &  36.4  &  38.2  \\ 
				& MCGZSL \cite{felix2018multi}  &  56.9 &  64.0 &  60.2  & 51.9  &  67.2 &  58.6  &  45.7  &  61.0 &  52.3  & 49.4  &  33.6  & 40.0 \\ 
				& VZSL \cite{wang2018zero}  &  53.4   & 68.3 & 59.9  &  51.7  &  67.2  & 58.4  &  44.9  &  54.1  & 49.1  & 43.5  &  34.9  &  38.7  \\  
				& ABPZSL \cite{zhu2019learning}   &  57.4  &  66.7  &  61.7  &  54.9  &  63.6  & 58.9  &  45.2  &  55.5  &  49.8     &  42.6  &   36.7   &  39.4 \\ 
				& CADA-VAE \cite{schonfeld2019generalized}  &  52.7  &  74.7  &  61.8  &  54.1   &  77.1  &  63.6  &  50.0  & 54.7       &  52.2  &  45.1  &  36.4   &  40.3  \\
				& Ours    &57.1     &68.1    & \textbf{62.1}    &56.7     &74.8    &\textbf{64.5}    &50.8    &54.9    & \textbf{52.8}    &46.7   &39.5    &\textbf{42.8}    \\ \bottomrule[2pt]
			\end{tabular}
			\vspace{-.1in}
	\end{table}}

	\section{Experiment}
	
	\subsection{Experiment Settings}
	
	\begin{wraptable}{r}{0.45\textwidth}
		\vspace{-.25in}
		\centering
		\caption{Zero-shot learning evaluation on AwA1, AwA2, CUB, and SUN datasets.}
		\vspace{.1in}
		\label{tab:zslResults}
		\scriptsize
		\renewcommand{\tabcolsep}{1.75mm}
		\begin{tabular}{c|l|cccc}
			\toprule[2pt]
			& \textbf{Methods} & AwA1 & AwA2 & CUB & SUN \\ \midrule[0.5pt]
			\multirow{7}{*}{\rotatebox{90}{\hspace{-1.3mm} Non-generative}} &
			IAP \cite{lampert2013attribute}    & 35.9 & 35.9 & 24.0& 19.4\\
			& CMT \cite{socher2013zero}    & 39.5 & 37.9 & 34.6& 39.9\\
			& LATEM \cite{xian2016latent}  & 55.1 & 55.8 & 49.3& 55.3\\
			& SYNC  \cite{changpinyo2016synthesized}  & 54.0 & 46.6 & 55.6& 56.3\\
			& ALE   \cite{akata2015label}  & 59.9 & 62.5 & 54.9& 58.1\\
			& ESZSL \cite{romera2015embarrassingly}  & 58.2 & 58.9 & 53.9& 54.5\\
			& DeViSE \cite{frome2013devise} & 54.2 & 59.7 & 52.0& 56.5\\ \midrule[0.5pt]
			\multirow{7}{*}{\rotatebox{90}{\hspace{-2.0mm} Generative}} &
			GAZSL \cite{zhu2018generative}  & 63.7 & 64.2 & 55.8& 60.1\\
			& FGZSL \cite{xian2018feature}  & 65.6 & 66.9 & 57.7& 58.6\\
			& MCGZSL \cite{felix2018multi} & 66.8 & 67.3 & 58.4& 60.0\\
			& VZSL  \cite{wang2018zero}  & 67.1 & 66.8 & 56.3& 59.0\\
			& ABPZSL \cite{zhu2019learning} & \textbf{69.7} & \textbf{70.3} & 56.2& 59.3\\
			& CADA-VAE \cite{schonfeld2019generalized} & 57.7 & 61.6 & \textbf{58.8} & 60.0\\
			& Ours & 66.4 & 66.6 & 57.2& \textbf{61.8}\\ \bottomrule[2pt]
		\end{tabular}
		\vspace{-.3in}
	\end{wraptable}
	
	\paragraph{Dataset.} We evaluate our method on five benchmark datasets, 
	including AwA1 (Animals with Attributes) \cite{lampert2013attribute}, 
	AwA2 (Animals with Attributes2) \cite{xian2017zero}, 
	CUB (Caltech-UCSDBirds-200-2011) \cite{wah2011caltech}, 
	SUN (SUN attributes) \cite{patterson2012sun},
	and ImageNet21K \cite{deng2009imagenet}.
	Among them, AwA1 and AwA2 are small-scale datasets with 10 unseen classes. 
	CUB and SUN are medium-scale datasets, containing 50 and 72 unseen classes, respectively. 
	ImageNet21K is usually used for large-scale ZSL/GZSL evaluation. 
	The features of all datasets are extracted using ResNet101 \cite{he2016deep} pretrained on the ImageNet1K dataset \cite{deng2009imagenet}.
	For a fair comparison, we follow the ZSL/GZSL evaluation protocol in \cite{xian2017zero}.
	
	\paragraph{Implementation.}
	The proposed VAE-cFlow is implemented in PyTorch \cite{paszke2019pytorch}. 
	The VAE encoder and decoder are multilayer perceptrons (MLP) with two hidden layers and one hidden layer, respectively. 
	The encoder has $1024$ and $2048$ hidden neurons, and the decoder has $512$ hidden neurons.
	Due to the invertibility of the generative flow, 
	the embedding space of VAE must match the dimension of the visual feature. Hence the VAE has $2048$ embedding dimensions. 
	We use four repeated invertible blocks for constructing the generative flow. 
	The building block is composed of the additive coupling layer and ActNorm. 
	We use shuffle permutation in the coupling layer to fuse different channels. 
	Our model is trained using the Adam \cite{kingma2014adam} optimizer with a learning rate $0.0001$.
	We use a batch size of $256$ for training ImageNet21K and $64$ for others.
	
	\subsection{Evaluation on Benchmark Datasets}
	
	The proposed VAE-cFlow is evaluated in both ZSL and GZSL settings. 
	We report average per-class top-1 accuracy for ZSL evaluation. 
	For GZSL evaluation, we follow the evaluation protocol in \cite{xian2017zero} to compute the harmonic mean of accuracy for seen and unseen classes: $H=\frac{2\cdot \mathit{A}_{\mathcal{U}}\cdot \mathit{A}_{\mathcal{S}}}{\mathit{A}_{\mathcal{U}}+\mathit{A}_{\mathcal{S}}}$, 
	where $\mathit{A}_{\mathcal{S}}$ and $\mathit{A}_{\mathcal{U}}$ represent the accuracy of seen and unseen classes, respectively. 
	We average ten runs to get the final results.
	
	\paragraph{Small- and Medium-scale ZSL/GZSL Evaluation.} 
	We compare the proposed VAE-cFlow with seven non-generative approaches and six generative approaches. 
	The performance of non-generative approaches is borrowed from the GBU benchmark \cite{xian2017zero}.  
	For generative approaches, we run the public code of FGZSL \cite{xian2018feature}, CADA-VAE \cite{schonfeld2019generalized} and ABPZSL \cite{zhu2019learning} to obtain their results. 
	The results of the other three generative methods are taken from  \cite{zhu2019learning}.
	The evaluation results in ZSL and GZSL settings are shown in \tabref{tab:zslResults} and \tabref{tab:GZSLresults}, respectively.
	The proposed VAE-cFlow has superior performance for GZSL, consistently improving previous generative methods on all datasets. Specifically,
	the improvement over CADA-VAE \cite{schonfeld2019generalized} is up to $2.5\%$ on the SUN dataset. 
	We visualize the synthetic and real unseen features using t-SNE \cite{maaten2008visualizing}.
	From \figref{fig:tsne} (a)-(b), we observe the synthetic features have clear inter-class separability and rich intra-class diversity as real ones. 
	Synthetic features with lower temperature $\mathcal{T}=0.5$ have less intra-class variations, as shown in \figref{fig:tsne} (c). 
	By joint visualization in \figref{fig:tsne} (d), we find the clusters of synthetic and real features are largely overlapped with each other, demonstrating the accurate estimation of the visual distribution. 
	
	\CheckRmv{
		\begin{figure}[t]
			\centering
			\subfloat[Real]{
				\includegraphics[width=0.24\linewidth]{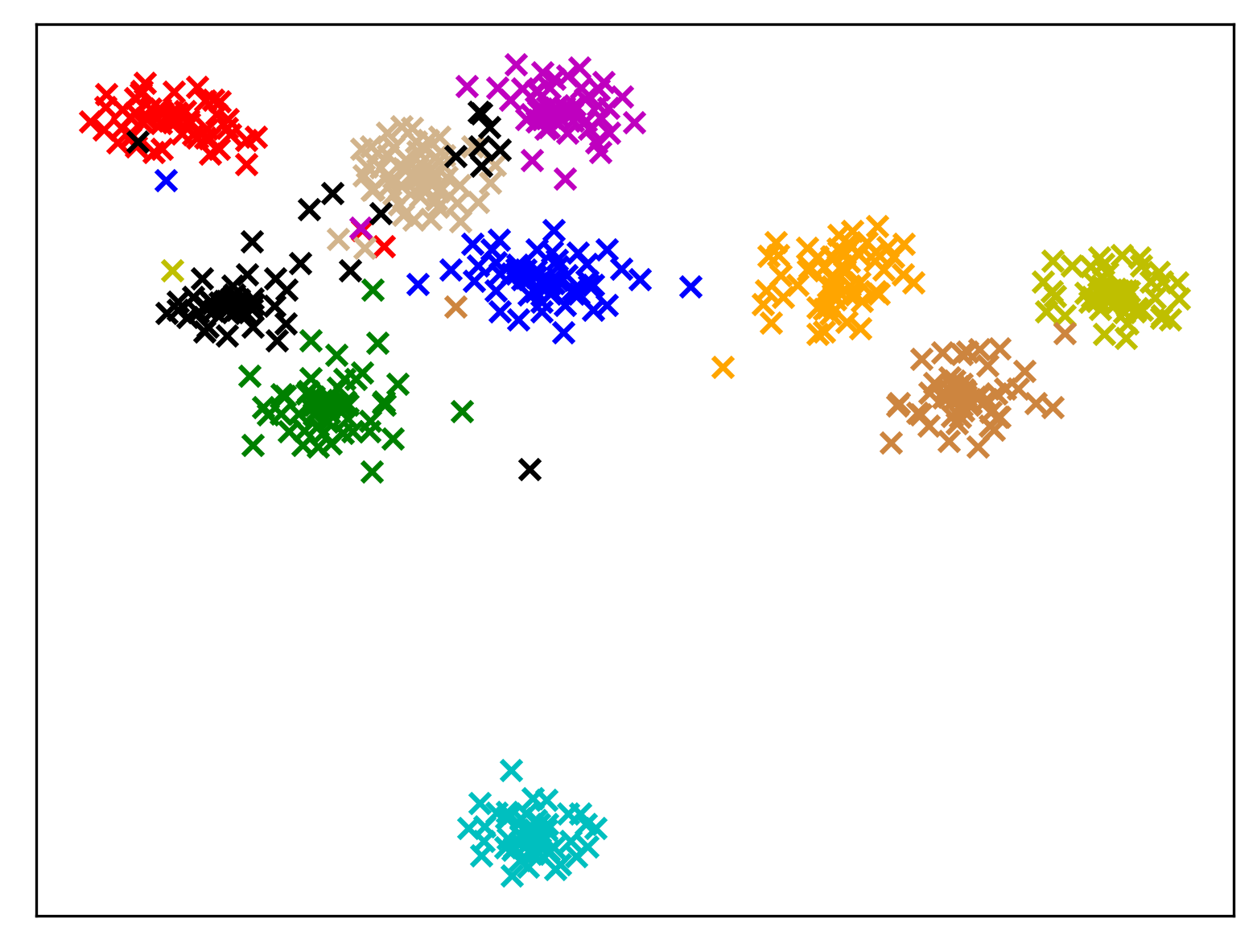}
			}
			\subfloat[Synthetic ($\mathcal{T}=1.0$)]{
				\includegraphics[width=0.24\linewidth]{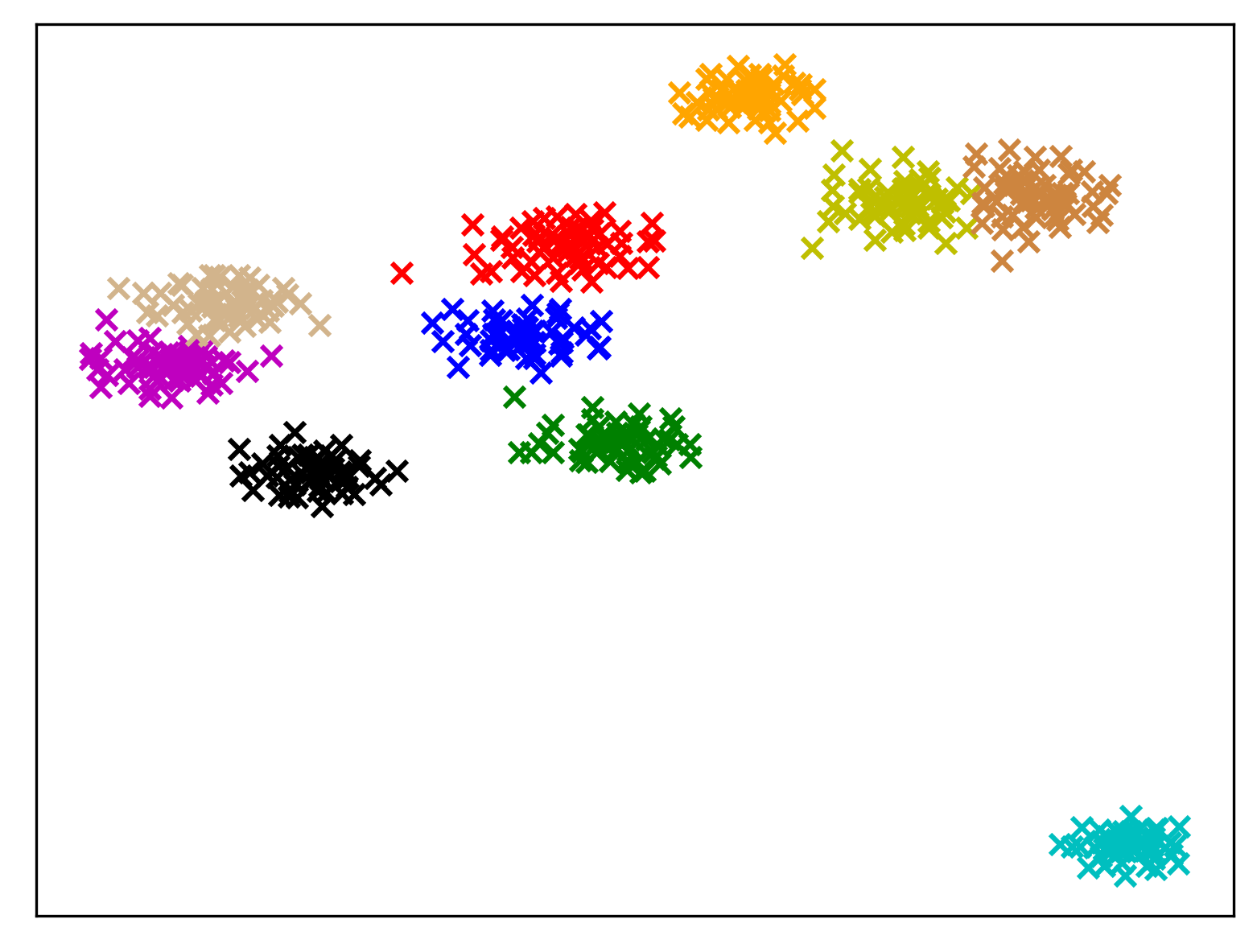}
			}
			\subfloat[Synthetic ($\mathcal{T}=0.5$)]{
				\includegraphics[width=0.24\linewidth]{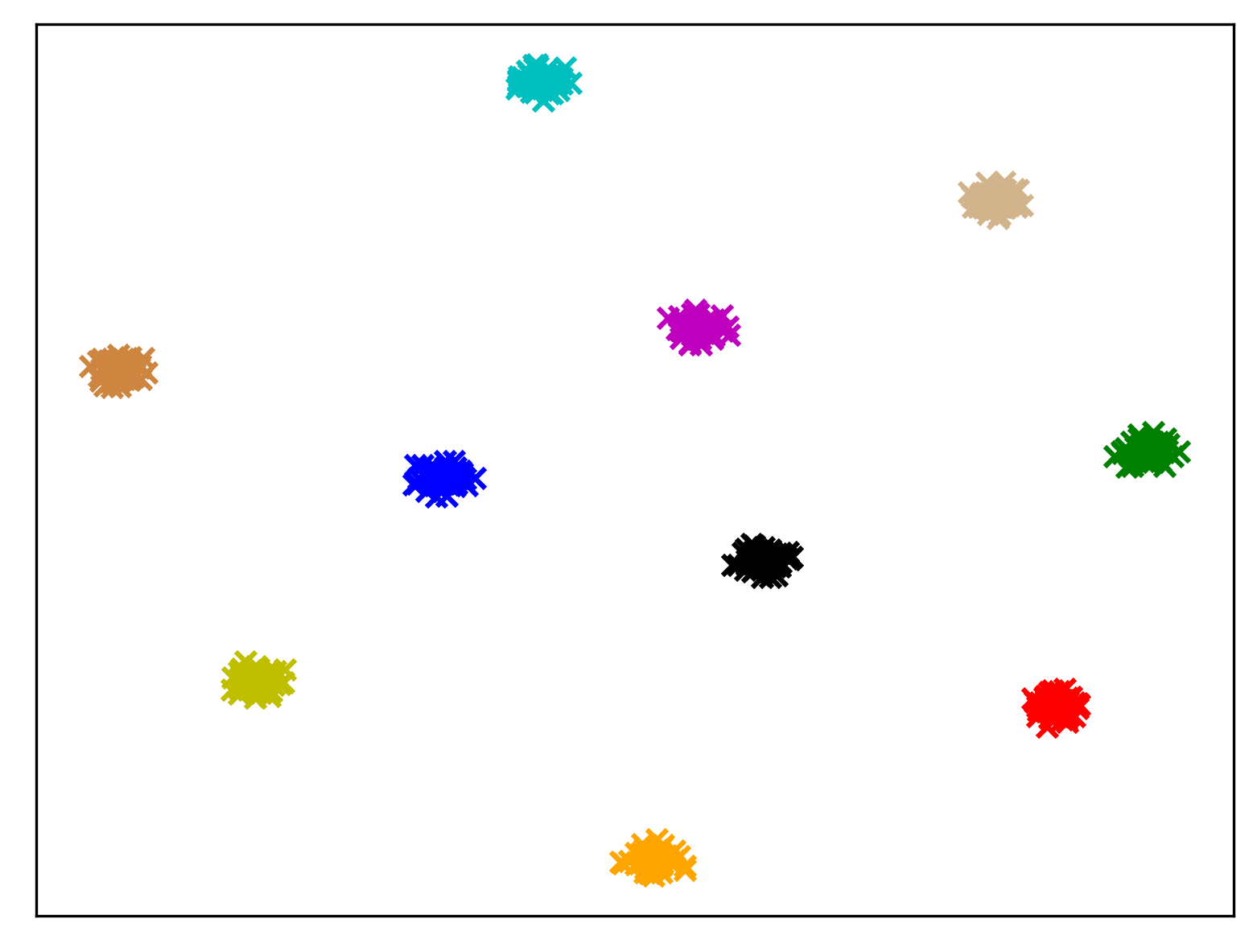}
			}
			\subfloat[Joint]{
				\includegraphics[width=0.24\linewidth]{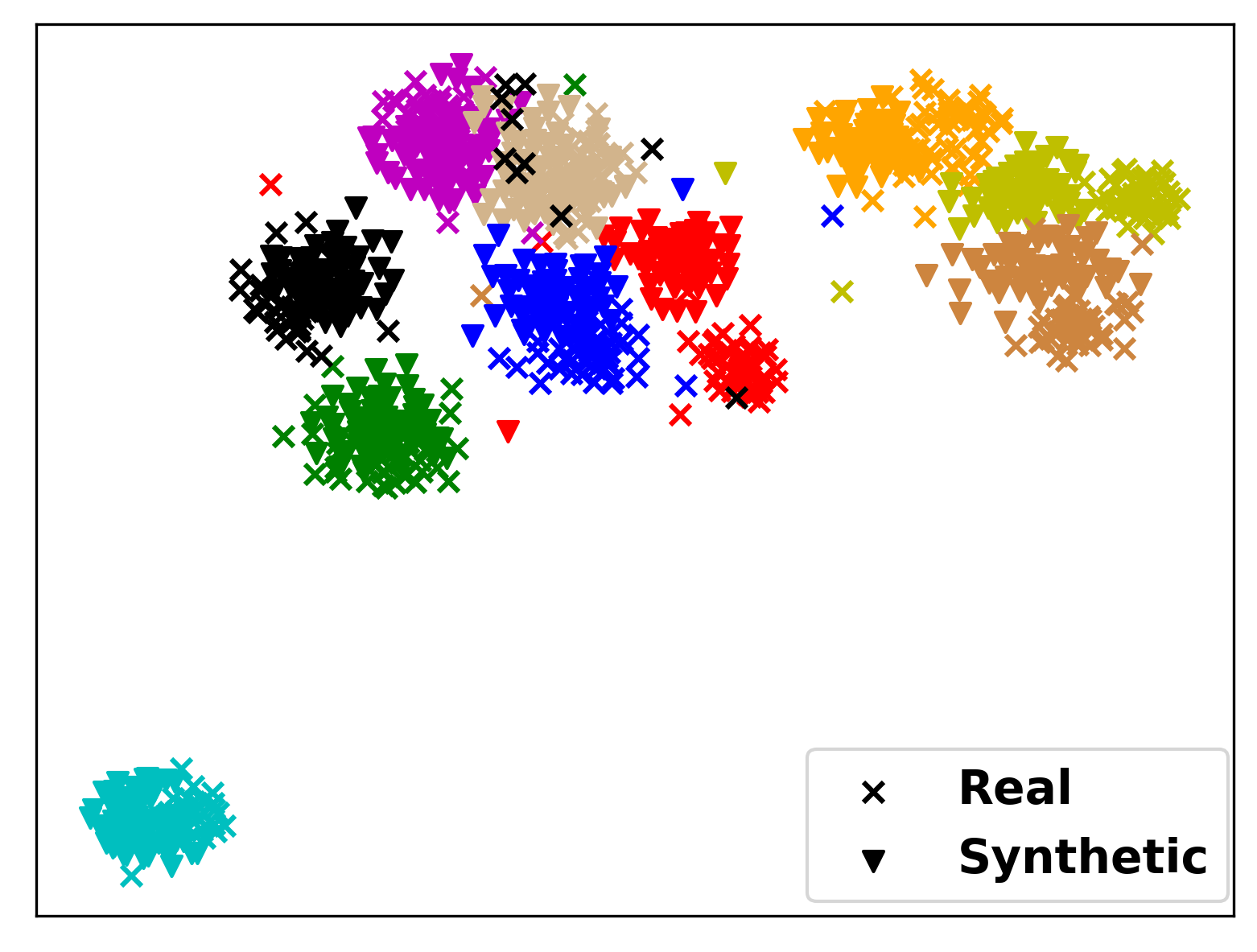}
			}
			\caption{T-SNE visualization of real and synthetic unseen features of the CUB dataset.
				(a) Real unseen features, (b) Synthetic unseen features with $\mathcal{T}=1.0$, 
				(c) Synthetic unseen features with $\mathcal{T}=0.5$, 
				(d) Joint visualization of real and synthetic features. 
				Zoom in for details.}
			\label{fig:tsne}
			\vspace{-.1in}
	\end{figure}}
	
	\CheckRmv{
		\begin{figure}[htb]
			\centering
			\includegraphics[width=.98\linewidth]{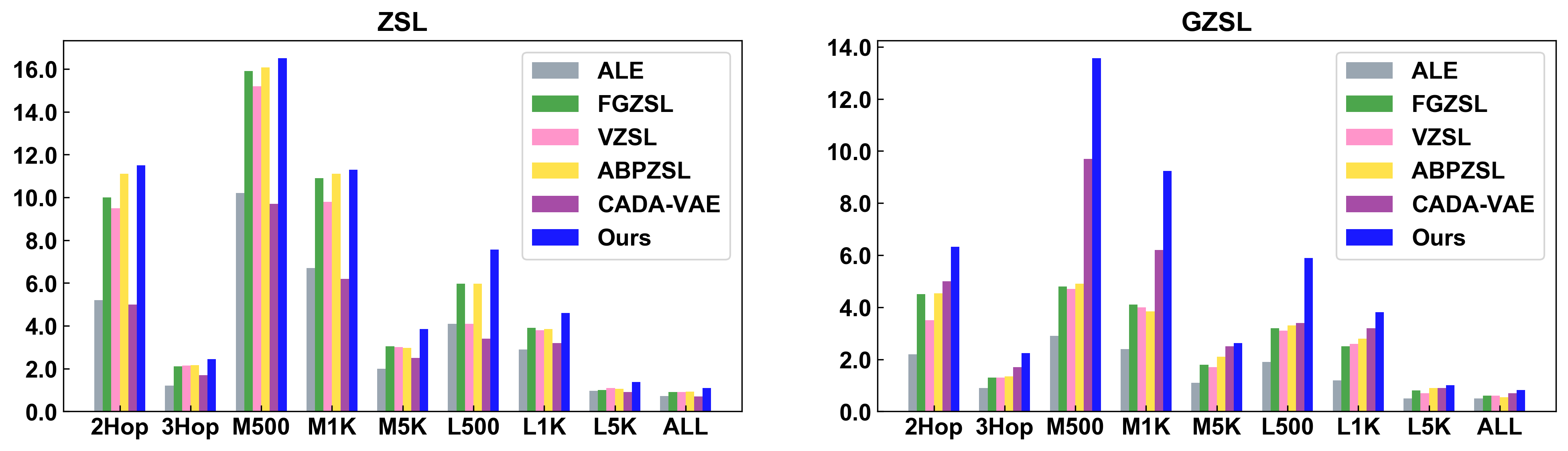} 
			\caption{Evaluation on the large-scale ImageNet21K dataset for ZSL and GZSL.}
			\label{fig:imagenet}
			\vspace{-.1in}
	\end{figure}}
	
	\paragraph{Large-scale ZSL/GZSL Evaluation.}
	For the evaluation on the large-scale ImageNet21K dataset \cite{deng2009imagenet}, we compare the proposed VAE-cFlow with one non-generative method, \textit{i.e.}, ALE \cite{akata2015label}, and four generative methods, including FGZSL \cite{xian2018feature}, VZSL \cite{wang2018zero}, ABPZSL \cite{zhu2019learning}, and CADA-VAE \cite{schonfeld2019generalized}.
	ImageNet21K has 1K seen classes and the remaining are unseen classes.
	The unseen classes are divided into several subsets based on different criteria.
	According to the hierarchical distance from 1K seen classes, the 2Hops split contains 1509 unseen classes and the 3Hops split contains 7678 classes. 
	According to the popularity of classes, the unseen classes are divided into the most popular 500 (M500), 1000 (M1K), 5000 (M5K) classes, 
	and the least popular 500 (L500), 1000 (L1K), 5000 (L5K) classes. 
	Finally, all unseen classes are used for evaluation (ALL).
	Evaluation results can be found in \figref{fig:imagenet}. 
	The proposed VAE-cFlow consistently improves previous generative methods in both ZSL and GZSL settings. 
	Especially, the improvement over CADA-VAE \cite{schonfeld2019generalized} on M500, M1K, and L500 in the GZSL setting is up to $3.9\%$, $3.1\%$, and $2.5\%$, respectively.

	\subsection{Ablation Study}
	
	\begin{wraptable}{r}{0.50\textwidth}
		\vspace{-.3in}
		\begin{center}
			\caption{Ablating different components of VAE-cFlow.}
			\vspace{.1in}
			\label{tab:ablation}
			\scriptsize
			\begin{tabular}{l|cc|cc}
				\toprule[2pt]
				\multirow{2}*{\textbf{Methods}} &  \multicolumn{2}{c|}{CUB} & \multicolumn{2}{c}{SUN} \\ 
				& ZSL      & GZSL      & ZSL      & GZSL   \\ \midrule[0.5pt]
				cFlow  &49.6    &44.9    &54.4    &37.1     \\
				VAE-cFlow w/o $\mathcal{L}_{cls}$   &53.2    &50.1    &57.5    &40.3     \\
				VAE-cFlow &\textbf{57.2}    &\textbf{52.8}    &\textbf{61.8}    &\textbf{42.8}     \\ \bottomrule[2pt]
			\end{tabular}
		\end{center}
		\vspace{-.2in}
	\end{wraptable}
	
	\paragraph{Effectiveness of Each Component.} 
	We verify the effectiveness of each design by ablating each component.
	We build the baseline cFlow from a conditional form of Glow \cite{kingma2018glow}, which exploits the linear layer to encode the parameters of the conditional latent distribution. 
	We replace the linear layer with VAE and release the zero-mean constraint in VAE posterior regularization, improving cFlow by $5.2\%$ and $3.6\%$ on the CUB dataset for GZSL and ZSL, respectively. 
	The improvement comes from that the conditional latent distribution encoded by VAE is more semantic meaningful, where the sampling latent can be reconstructed to semantic embedding. 
	To make the latent distribution inter-class discriminative, we add  $\mathcal{L}_{cls}$ (\eqref{eq:loss_cls}) on the latent variable encoded by the generative flow.
	This classification regularization further improves GZSL by $2.7\%$ on the CUB and $2.5\%$ on the SUN dataset.
	
	\begin{wrapfigure}{r}{0.4\textwidth}
		\vspace{-.45in}
		\begin{center}
			\includegraphics[width=0.38\textwidth,height=0.22\textwidth]{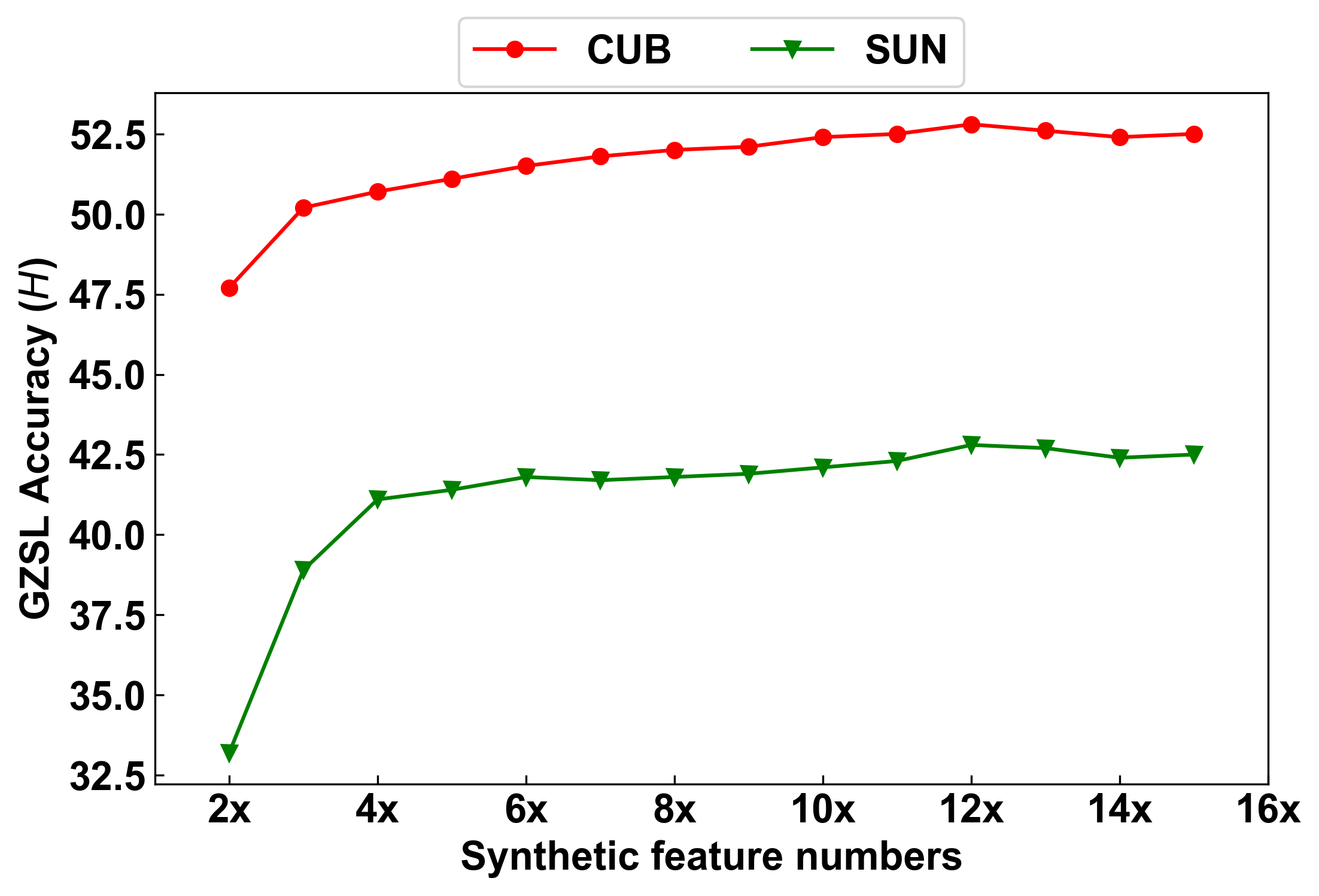}
		\end{center}
		\vspace{-.15in}
		\caption{Comparison of GZSL performance with different synthetic feature numbers.
		}
		\vspace{-.3in}
		\label{fig:number}
	\end{wrapfigure}
	
	\paragraph{Number of Synthetic Unseen Features.}
	\figref{fig:number} shows the GZSL performance with different numbers of synthetic unseen features per class, ranging from $1\times$ and $15\times$ of seen features. 
	We observe significant performance improvement when the number increases from $2\times$ to $4\times$. The performance is improved steadily with the increasing of synthetic unseen features till the number reaches $12\times$ of seen features. 
	Without other mentions, we generate $12\times$ unseen features than
	seen features in this work. 
	
	\begin{wrapfigure}{r}{0.5\textwidth}
		\vspace{-.2in}
		\begin{center}
			\includegraphics[width=0.48\textwidth,height=0.28\textwidth]{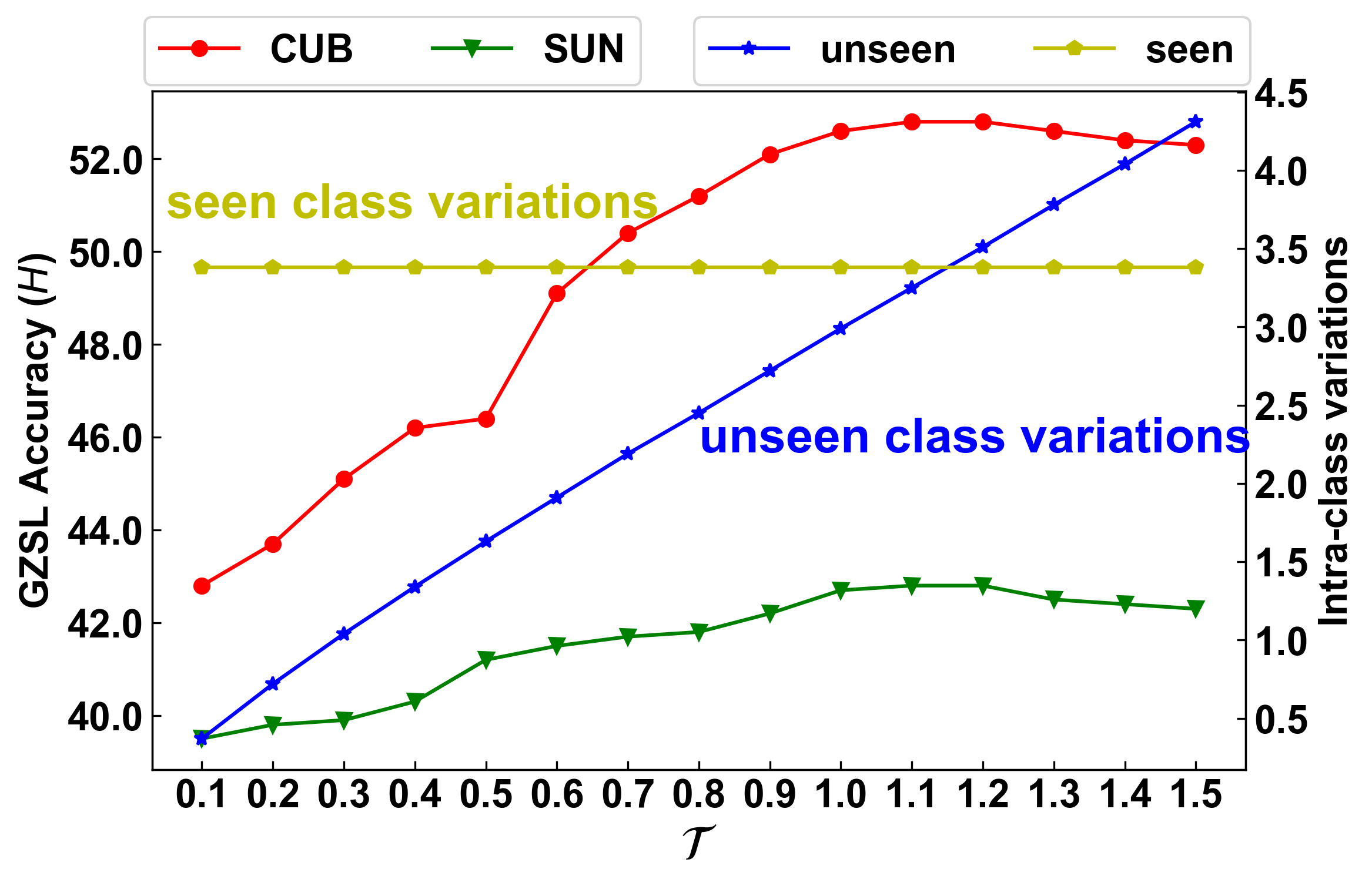}
		\end{center}
		\vspace{-.15in}
		\caption{Comparison of GZSL performance with different sampling temperature $\mathcal{T}$.}
		\vspace{-.15in}
		\label{fig:var}
	\end{wrapfigure}
	
	\paragraph{Influence of Temperature.}
	We conduct experiments to evaluate the effect of temperature $\mathcal{T}$ in \eqref{eq:test}. 
	In the testing phase, $\mathcal{T}$ controls the variation of the latent distribution and thus controls the intra-class variations of the synthetic features, as shown in \figref{fig:var}. \figref{fig:tsne} (b) and (c) show the t-SNE visualization of synthetic features under $\mathcal{T}=1.0$ and $\mathcal{T}=0.5$. Small $\mathcal{T}$ values mean more intra-class compactness. When the intra-class variation of synthetic unseen features matches that of real seen features, \textit{i.e.}, $\mathcal{T}=1.1$, GZSL achieves the best performance.
	
	\subsection{Discussion}
	Overall, our work establishes a new generative method VAE-cFlow for GZSL. 
	The key to our approach is to construct meaningful conditional space and optimize the accurate log-likelihood objective. 
	We empirically find that the variation of synthetic unseen features is crucial to GZSL. 
	When the variation of synthetic unseen features matches that of seen classes, GZSL achieves the best performance. 
	This finding is also in line with previous results. 
	More specifically, non-generative methods consider each semantic class as a deterministic vector, ignoring the variation of unseen classes. 
	Hence non-generative methods usually have poor performance in the GZSL setting. The proposed VAE-cFlow shows superiority in ZSL/GZSL on the large-scale setting, but inferiority in ZSL with the small-scale setting, which is a limitation of VAE-cFlow.
	Our work opens a new perspective for future research on GZSL by constructing invertible models to accurately estimate the conditional visual distribution.
	
	\section{Conclusion}
	We have introduced VAE-cFlow, a conditional generative approach for generalized zero-shot learning. 
	Different from previous GAN- and VAE-based generative methods, we resort to the generative flow, a more stable and theoretically accurate method. 	
	The proposed VAE-cFlow encodes the semantic description into a conditional latent distribution through VAE. Conditioned on that, the generative flow optimizes the exact log-likelihood of the observed visual features. We ensure the conditional latent distribution to be both semantic meaningful and inter-class discriminative by i) 
	adopting the VAE reconstruction objective, ii) releasing the zero-mean constraint in the VAE posterior regularization, and iii) adding a classification regularization on the latent variable.
	We experimentally demonstrate that the realistic variation of unseen classes modeled by generative methods is the key to achieve better GZSL performance. 
	The VAE-cFlow outperforms previous generative methods on five GZSL benchmark datasets, especially in the large-scale setting. 
	The visualization through t-SNE demonstrates that our synthetic features are more consistent with the real ones.
	
	\section{Broader Impact Discussion}
	This paper proposes the VAE-Conditioned Generative Flow (VAE-cFlow) for generalized zero-shot learning.
	For artificial intelligence practitioners, this work shows a possible solution to save human efforts by collecting and annotating data of a few classes but training models for the recognition of more classes.
	At last, this work has no obvious harm to society. However, this work can also be a tool for constructing large-scale visual recognition systems as other computer vision techniques. It is not our original intention if these recognition systems are used for some bad purposes.
	
	\bibliographystyle{plain}
	\bibliography{VAEcFlow}
	
	\newpage

	\section{More Experimental Details and Results}
    In this document, we clarify detailed experimental settings for the classifier (\secref{sec:classifier}) and hyperparameters (\secref{sec:hyper}). We also provide more experimental results of \textit{zero-shot learning} (ZSL) and \textit{generalized zero-shot learning} (GZSL) on the ImageNet dataset \cite{deng2009imagenet} (\secref{sec:imagenet}).

    \subsection{Classifier Details} \label{sec:classifier}
    After obtaining sufficient features for unseen classes by VAE-cFlow, we view the problem as a classification task to evaluate ZSL and GZSL. Following previous works \cite{xian2018feature,zhu2019learning}, we train a simple linear classifier with the \textit{softmax} activation. The Adam optimizer \cite{kingma2014adam} is used for training with a learning rate of 1$e$-3, $\beta_1$ of 0.5, and $\beta_2$ of 0.999. We train the classifier for 20 epochs in all experiments.

    \subsection{Hyperparameters of VAE-cFlow} \label{sec:hyper}
    The hyperparameters for training VAE-cFlow on different datasets are given in \tabref{tab:hyper}. The number of training iterations increases with the number of seen classes, because it is widely accepted that more data usually need more training iterations. The number of synthetic unseen features is approximately 12$\times$ more than the real seen features. It is worth noting that due to the large number of real features and unseen classes for the ImageNet dataset \cite{deng2009imagenet}, we randomly choose 50 features for each seen class and generate 600 features for each unseen class to train a \textit{softmax} classifier. The batch size for the ImageNet dataset \cite{deng2009imagenet} is also larger owing to the large scale of data.
    
    \CheckRmv{
    \begin{table}[ht]
    \centering
    \caption{Hyperparameters for training VAE-cFlow on different datasets. $\mathcal{Y}_s$ and $\mathcal{Y}_u$ mean the numbers of seen classes and unseen classes, respectively. The symbol $a$ means the dimension of semantic descriptions. The symbols of $iter$ and $bs$ mean the number of training iterations and the batch size for training VAE-cFlow. $\#$Synthetic and $\#$Real are the numbers of synthetic unseen features and real seen features for each class, respectively.}
    \label{tab:hyper}
    \small
    \begin{tabular}{lcccccccc}
    \toprule[2pt]
    Datasets & Size & $\mathcal{Y}_s$ & $\mathcal{Y}_u$ & $a$ & $iter$  & $bs$ & $\#$Synthetic  & $\#$Real \\ \midrule[1pt]
    AWA1  &  Small & 40   & 10 &      85 &   5K    &    64      &      6000   & 500    \\
    AWA2  &  Small & 40   & 10 &      85 &   5K    &    64      &     6000   & 500    \\
    CUB   &  Medium & 150 & 50 &   312 &   25K   &    64      &     600   & 50  \\
    SUN   &  Medium & 645 & 72 &  102  &    28K  &    64      & 192  & 16  \\
    ImageNet & Large & 1000  & 20345  &  500 &  150K   &    256      &  600   & 50    \\ \bottomrule[2pt]
    \end{tabular}
    \end{table}}

    \subsection{More Experimental Results on ImageNet} \label{sec:imagenet}
    We provide more experimental results for ZSL and GZSL on the ImageNet dataset \cite{deng2009imagenet} in \tabref{tab:zsl} and \tabref{tab:gzsl}, respectively. We find that generative methods have significant superiority over non-generative methods in both ZSL and GZSL settings. Among generative methods, ABPZSL \cite{zhu2019learning} achieves good ZSL performance but unsatisfactory GZSL performance. Moreover, CADA-VAE \cite{schonfeld2019generalized} works well for large-scale GZSL but lays behind in the ZSL setting. Our proposed VAE-cFlow improves both the ZSL and GZSL performance. Specially, the improvement over CADA-VAE on the M500, M1K, and L500 splits in the GZSL setting is up to $3.9\%$, $3.1\%$, and $2.5\%$, respectively. Since the number of ImageNet classes is large, such improvement is rather significant.
    
    \CheckRmv{
    \begin{table}[ht]
    \centering
    \caption{Zero-shot learning evaluation in terms of top-1 accuracy on the different splits of the ImageNet dataset \cite{deng2009imagenet}.}
    \label{tab:zsl}
    \begin{tabular}{c|lcc|ccc|ccc|c}
    \toprule[2pt]
     && \multicolumn{2}{c}{\textbf{Hierarchy}} & \multicolumn{3}{c}{\textbf{Most Populated}} & \multicolumn{3}{c}{\textbf{Least Populated}} & \textbf{ALL} \\ 
    & Method                        & 2H             & \multicolumn{1}{c}{3H}            & 500         & 1K        & \multicolumn{1}{c}{5K}        & 500         & 1K         & \multicolumn{1}{c}{5K}        & 20K \\ \midrule[1pt]
    \multirow{5}{*}{\rotatebox{90}{\hspace{-1.3mm} \small{Non-generative}}} & CMT \cite{socher2013zero} & 2.9 & 0.7 & 5.1 & 3.0 & 1.0 & 1.9 & 1.1 & 0.3 & 0.3 \\
    & LATEM \cite{xian2016latent} & 5.5 & 1.3 & 10.8 & 6.6 & 1.9 & 4.5 & 2.7 & 0.8 & 0.5 \\
    & ALE \cite{akata2015label} & 5.4 & 1.3 & 10.4 & 6.8 & 2.0 & 4.3 & 2.9 & 0.8 & 0.5 \\
    & ESZSL \cite{romera2015embarrassingly} & 6.4 & 1.5 & 11.9 & 7.7 & 2.3 & 4.5 & 3.2 & 0.9 & 0.6 \\
    & DeViSE \cite{frome2013devise} & 5.3 & 1.3 & 10.4 & 6.7 & 1.9 & 4.2 & 2.9 & 0.8 & 0.5 \\ \midrule[0.5pt] 
    \multirow{5}{*}{\rotatebox{90}{\hspace{-2.0mm} \small{Generative}}} & FGZSL \cite{xian2018feature} & 10.0 & 2.1 & 15.9 & 10.9 & 3.1 & 6.0 & 3.9 & 1.0 & 0.9 \\
    & VZSL \cite{wang2018zero} & 9.5 & 2.2 & 15.2 & 9.8 & 3.0 & 4.1 & 3.8 & 1.1 & 0.9 \\ 
    & ABPZSL \cite{zhu2019learning} & 11.1 & 2.2 & 16.1 & 11.1 & 3.0 & 6.0 & 3.9 & 1.1 & 0.9 \\
    & CADA-VAE \cite{schonfeld2019generalized} & 5.0 & 1.7 & 9.7 & 6.2 & 2.5 & 3.4 & 3.2 & 0.9 & 0.7 \\ 
    & Ours & \textbf{11.5} & \textbf{2.5} & \textbf{16.5} & \textbf{11.3} & \textbf{3.9} & \textbf{7.6} & \textbf{4.6} & \textbf{1.4} & \textbf{1.1} \\ \bottomrule[2pt]
    \end{tabular}
    \end{table}}
    
    \CheckRmv{
    \begin{table}[ht]
    \centering
    \caption{Generalized zero-shot learning evaluation on the different splits of the ImageNet dataset \cite{deng2009imagenet}. The metric is the harmonic mean of top-1 accuracy for seen and unseen classes.}
    \label{tab:gzsl}
    \begin{tabular}{c|lcc|ccc|ccc|c}
    \toprule[2pt]
     && \multicolumn{2}{c}{\textbf{Hierarchy}} & \multicolumn{3}{c}{\textbf{Most Populated}} & \multicolumn{3}{c}{\textbf{Least Populated}} & \textbf{ALL} \\ 
    & Method                        & 2H             & \multicolumn{1}{c}{3H}            & 500         & 1K        & \multicolumn{1}{c}{5K}        & 500         & 1K         & \multicolumn{1}{c}{5K}        & 20K \\ \midrule[1pt]
    \multirow{5}{*}{\rotatebox{90}{\hspace{-1.3mm} \small{Non-generative}}} & CMT \cite{socher2013zero} & 1.1 & 0.4 & 1.6 & 1.2 & 0.5 & 0.8 & 0.4 & 0.2 & 0.2 \\
    & LATEM \cite{xian2016latent} & 2.0 & 0.7 & 2.6 & 2.1 & 1.0 & 1.0 & 0.6 & 0.4 & 0.4\\
    & ALE \cite{akata2015label} & 2.2 & 0.9 & 2.9 & 2.4 & 1.1 & 1.9 & 1.2 & 0.5 & 0.5\\
    & ESZSL \cite{romera2015embarrassingly} & 1.4 & 0.5 & 1.6 & 1.5 & 0.7 & 0.6 & 0.4 & 0.3 & 0.3\\
    & DeViSE \cite{frome2013devise} & 2.1 & 0.8 & 2.9 & 2.3 & 1.1 & 1.6 & 1.3 & 0.4 & 0.3\\ \midrule[0.5pt] 
    \multirow{5}{*}{\rotatebox{90}{\hspace{-1.3mm} \small{Generative}}} & FGZSL \cite{xian2018feature} & 4.5 & 1.3 & 4.8 & 4.1 & 1.8 & 3.2 & 2.5 & 0.8 & 0.6\\
    & VZSL \cite{wang2018zero} & 3.5 & 1.3 & 4.7 & 4.0 & 1.7 & 3.1 & 2.6 & 0.7 & 0.6\\ 
    & ABPZSL \cite{zhu2019learning} & 4.5 & 1.4 & 4.9 & 3.9 & 2.1 & 3.3 & 2.8 & 0.9 & 0.6 \\
    & CADA-VAE \cite{schonfeld2019generalized} & 5.0 & 1.7 & 9.7 & 6.2 & 2.5 & 3.4 & 3.2 & 0.9 & 0.7\\ 
    & Ours & \textbf{6.3} & \textbf{2.2} & \textbf{13.6} & \textbf{9.3} & \textbf{2.6} & \textbf{5.9} & \textbf{3.8} & \textbf{1.0} & \textbf{0.8}\\ \bottomrule[2pt]
    \end{tabular}
    \end{table}}

\end{document}